\documentclass[11pt,letterpaper]{article}
\usepackage{emnlp2017}
\usepackage{times}
\usepackage{latexsym}
\usepackage{amsmath}
\usepackage{amssymb}

\usepackage{graphicx}
\graphicspath{ {images/} }

\emnlpfinalcopy

\def\emnlppaperid{***}

\DeclareMathOperator*{\argmax}{arg\,max}

\title{Does Neural Machine Translation Benefit from Larger Context?}

\author{Sebastien Jean\textbf{*} \and Stanislas Lauly\textbf{*} \and Orhan Firat\thanks{~~~This work was done during his visit to NYU. Now at Google (orhanf@google.com).}  \and Kyunghyun Cho \\
        Department of Computer Science \\ 
        Center for Data Science \\ 
        New York University\\
        \textbf{*} \textit{Both authors contributed equally}}

\date{}

\begin{document}

\maketitle

\begin{abstract}

We propose a neural machine translation architecture that models the surrounding text in addition to the source sentence. These models lead to better performance, both in terms of general translation quality and pronoun prediction, when trained on small corpora, although this improvement largely disappears when trained with a larger corpus. We also discover that attention-based neural machine translation is well suited for pronoun prediction and compares favorably with other approaches that were specifically designed for this task.

\end{abstract}

\section{Introduction}

A major strength of neural machine translation, which has recently become {\it de facto} standard in machine translation research, is the capability of seamlessly integrating information from multiple sources. Due to the nature of continuous representation used within a neural machine translation system, any information, in addition to tokens from source and target sentences, can be integrated as long as such information can be projected into a vector space. This has allowed researchers to build a non-standard translation system, such as multilingual neural translation systems~\citep[see, e.g.,][]{firat2016multi,zoph2016multi}, multimodal translation systems~\citep[see, e.g.,][]{caglayan2016multimodal,specia2016shared} and syntax-aware neural translation systems~\citep[see, e.g.,][]{nadejde2017syntax,eriguchi2016tree,eriguchi2017learning}. At the core of all these recent extensions is the idea of using context larger than a current source sentence to facilitate the process of translation.

In this paper, we try our first attempt at investigating the potential for implicitly incorporating discourse-level structure into neural machine translation. As an initial attempt, we focus on incorporating a small number of preceding and/or following source sentences into the attention-based neural machine translation model~\citep{bahdanau2014neural}. More specifically, instead of modelling the conditional distribution $p(Y|X)$ over translations given a source sentence, we build a network that models the conditional distribution $p(Y|X, X_{-n}, \ldots, X_{-1}, X_{1}, \ldots, X_{n})$, where $X_{-i}$ is the $i$-th preceding source sentence, and $X_i$ the $i$-th following source sentence. We propose a novel larger-context neural machine translation model based on the recent works on larger-context language modelling~\citep{wang2016larger} and multi-way, multilingual neural machine translation~\citep{firat2016multi}.

We first evaluate the proposed model against the baseline model without any context other than a source sentence using BLEU and RIBES~\citep{isozaki2010automatic}, both of which measure translation quality {\it averaged} over all the sentences in a corpus. This evaluation strategy reveals that the benefit of larger context is not always apparent when the evaluation metric is average translation quality, confirming the earlier observation, for instance, by \citet{hardmeier-EtAl:2015:DiscoMT}. Then, we turn to a more focused evaluation based on pronoun prediction~\citep{guillou-EtAl:2016:WMT} which was a shared task at WMT'16. On this cross-lingual pronoun prediction task, we notice benefits from incorporating larger context when training models on small corpora, but not on larger ones. Interestingly, we also observe that neural machine translation can predict pronouns as well as the top ranking approaches from the shared task at WMT'16.

\section{Larger-Context \\ Neural Machine Translation}

\subsection{Attention-based \\ Neural Machine Translation}

Attention-based neural machine translation, proposed by \citet{bahdanau2014neural}, has become {\it de facto} standard in recent years, both in academia~\citep{bojar-EtAl:2016:WMT1} and industry~\citep{wu2016google,crego2016systran}. An attention-based translation system consists of three components; (1) encoder, (2) decoder and (3) attention model. The encoder is often a bidirectional recurrent network with a gated recurrent unit~\cite[GRU,][]{cho2014learning,hochreiter1997long}, which encodes a source sentence $X=(x_1, x_2, \ldots, x_{T_x})$ into a set of annotation vectors $\left\{ h_1, h_2, \ldots, h_{T_x} \right\}$, where $h_t = \left[\overrightarrow{h}_t; \overleftarrow{h}_t\right]$. $\overrightarrow{h}_t$ and $\overleftarrow{h}_t$ are the $t$-th hidden states from the forward and reverse recurrent networks respectively. 

The decoder is a recurrent language model~\citep{mikolov2010recurrent,Graves13} which generates one target symbol $y_{t'}$ at a time by first computing the attention scores $\left\{ \alpha_{t,t'} \right\}_{t=1}^{T_x}$ over the annotation vectors. Each attention score is computed by
\begin{align*}
    \alpha_{t,t'} \propto \exp(f_{\text{att}}(\hat{y}_{t'-1}, z_{t'-1}, h_t)),
\end{align*}
where $f_{\text{att}}$ is the attention model implemented as a feedforward network taking as input the previous target symbol $\hat{y}_{t'-1}$, the previous decoder hidden state $z_{t'-1}$ and one of the annotation vector $h_t$. These attention scores are used to compute the time-dependent source vector $s_{t'}=\sum_{t=1}^{T_x} \alpha_{t,t'} h_t$, based on which the decoder's hidden state and the output distribution over all possible target symbols are computed:
\begin{align*}
    p(y_{t'}|y_{<t'}, X) \propto \exp(g^{y_{t'}}_{\text{out}}(z_{t'})),
\end{align*}
where
\begin{align}
    \label{eq:dec_rnn}
    z_{t'} = \phi(\hat{y}_{t'-1}, z_{t'-1}, s_{t'}).
\end{align}
$\phi$ is a recurrent activation function such as a GRU or long short-term memory (LSTM) unit.

The whole model, consisting of the encoder, decoder and attention model, is fully differentiable, and can be jointly trained by maximizing the log-likelihood given a training corpus using stochastic gradient descent with backpropagation-through-time~\citep{werbos1990backpropagation}.

\subsection{Larger-Context \\ Neural Machine Translation}

We extend the attention-based neural machine translation described above by including an additional set of an encoder and attention model. This additional encoder is similarly a bidirectional recurrent network, and it encodes a context sentence, in our case a source sentence immediately before the current source sentence,\footnote{
Although we use a single preceding sentence in this paper, the proposed method can easily handle multiple preceding and/or following sentences either by having multiple sets of encoder and attention mechanism or by concatenating all the context sentences into a long single sequence. 
}
into a set of {\it context annotation vectors} $\left\{ h^c_1, \ldots, h^c_{T_c}\right\}$, where $h^c_t = \left[ \overrightarrow{h}^c_t; \overleftarrow{h}^c_t\right]$. Similarly to the original source encoder, these two vectors are from the forward and reverse recurrent networks. 

On the other hand, the additional attention model is different from the original one. The goal of incorporating larger context into translation is to provide additional discourse-level information necessary for translating a given source token, or a phrase. This implies that the attention over, or selection of, tokens from larger context be done with respect to which source token, or phrase, is being considered. We thus propose to make this attention model take as input the previous target symbol, the previous decoder hidden state, a context annotation vector as well as the source vector from the main attention model. That is,
\begin{align*}
    \alpha^c_{t,t'} \propto \exp(f^c_{\text{att}}(\hat{y}_{t'-1}, z_{t'-1}, h^c_t, s_{t'})).
\end{align*}
Similarly to the source vector, we compute the time-dependent {\it context vector} as the weight sum of the context annotation vectors: $c_{t'} = \sum_{t=1}^{T_c} \alpha^c_{t,t'} h^c_t$. 

Now that there are two vectors from both the current source sentence and the context sentence, the decoder transition in Eq.~\eqref{eq:dec_rnn} changes accordingly:
\begin{align}
    \label{eq:ls_dec_rnn}
    z_{t'} = \phi(\hat{y}_{t'-1}, z_{t'-1}, s_{t'}, c_{t'}).
\end{align}

We call this model a {\it larger-context neural machine translation} model. 



\section{Evaluating Larger-Context \\ Neural Machine Translation}

A standard metric for automatically evaluating the translation quality of a machine translation system is BLEU~\citep{papineni2002bleu}. BLEU is computed on a validation or test corpus by inspecting the overlap of $n$-grams (often up to 4-grams) between the reference and generated corpora. BLEU has become {\it de facto} standard after it has been found to correlate well with human judgement for phrase-based and neural machine translation systems. Other metrics, such as METEOR~\cite{denkowski:lavie:meteor-wmt:2014} and TER~\cite{snover2006study}, are often used together with BLEU, and they also measure the {\it average translation quality} of a machine translation system over an entire validation or test corpus. 

It is not well-known how much positive or negative effect larger context has on machine translation. It is understood that larger context allows a machine translation system to capture properties not apparent from a single source sentence, such as style, genre, topical patterns, discourse coherence and anaphora \citep[see, e.g., the preface of][]{DiscoMT:2015}, but the degree of its impact on the average translation quality is unknown. 

It is rather agreed that the impact should be measured by a metric specifically designed to evaluate a specific effect of larger context. For instance, discourse coherence has been used as one of such metrics in analyzing larger-context language modelling in recent years~\citep{ji2015document,ji2016latent}. In the context of machine translation, cross-lingual pronoun prediction~\citep{hardmeier-EtAl:2015:DiscoMT,guillou2016findings} has been one of the few established tasks by which the effect of larger-context modelling, or the ability of a machine translation system for incorporating larger-context information, is evaluated.

In this paper, we therefore compare the vanilla neural machine translation model against the proposed larger-context model based on both the average translation quality, measured by BLEU, {\it and} the pronoun prediction accuracy, measured in macro-averaged recall. In order to further investigate the relationship between the average translation quality and the pronoun prediction accuracy, we use a single corpus per language pair provided as a part of the 2016 WMT shared task on cross-lingual pronoun prediction~\citep{guillou2016findings}. 

Unlike the existing approaches to cross-lingual pronoun prediction, we do not train any of the models specifically for the pronoun prediction task, but train them to maximize the average translation quality. Once the model is trained, we conduct pronoun prediction by 
\begin{align}
\label{eq:pronoun_pred}
    \hat{y} = \argmax_{y \in P} \log p(y_{<n}^*,y,y_{> n}^*|X),
\end{align}
where $P$ is the set of all possible pronouns,\footnote{
In addition all possible pronouns, there is a class designated for any non-pronoun token.
} 
and the goal is to predict the pronoun in the $n$-th position in the target sentence.

\section{Experimental Settings}
\label{sec:experiments}

\subsection{Data and Tasks}

We use En-Fr and En-De for our experiments. The target side of the parallel corpus for each language pair has been heavily preprocssed, including tokenization and lemmatization. Although both of the corpora come with POS tags, we do not use them. In the case of En-Fr, the set $P$ of all pronouns includes ``ce'', ``elle'', ``elles'', ``il'', ``ils'', ``cela'', ``on'' and OTHER. The set consists of ``er'', ``sie'', ``es'', ``man'' and OTHER in the case of En-De. Macro-average recall is used as a main evaluation metric. There are 2,441,410 and 2,356,313 sentence pairs in the En-Fr and En-De training corpora, respectively.

For pronoun prediction, the input to the model is a source sentence and the corresponding target sentence of which some pronouns are replaced with a special token REPLACE. The goal is then to figure out which pronoun should replaced the REPLACE token, and this is done by finding a combination that maximizes the log-probability, as in Eq.~\eqref{eq:pronoun_pred}. When there are multiple REPLACE tokens in a single example, we exhaustively try all possible combinations, which is feasible as the size of the pronoun set $P$ is small.

For translation, the input to the model is a source sentence alone, and the model is expected to generate a translation. We use beam search to approximately find the maximum-a-posterior translation, i.e, $\argmax_Y \log p(Y|X)$. 

In addition to the data/tasks from the cross-lingual pronoun prediction shared task, we also check the average translation quality using IWSLT'15 En-De as training set. We use the IWSLT'12 and IWSLT'14 test set for development and test respectively. This is to ensure that our observation from the earlier lemmatized corpora transfers to non-lemmatized ones. This corpus has 194,371 sentence pairs for training, and 1700 and 1305 for development and test. 

\begin{table*}
\small
\begin{minipage}[t]{0.49\textwidth}
\centering
\begin{tabular}{l c c c c c}
& 5\% & 10\% & 20\% & 40\% & 100\% \\
\hline\hline
& \multicolumn{5}{c}{En-Fr} \\
\hline
NMT & 27.6 & 32.7 & 35.7 & 38.2 & 39.9 \\
LC-NMT & 28.8 & 33.9 & 36.7 & 38.6 & 39.0 \\
\hline\hline
& \multicolumn{5}{c}{En-De} \\
\hline
NMT & 16.3 & 19.8 & 22.1 & 24.3 & 25.6 \\
LC-NMT & 17.4 & 20.9 & 22.7 & 23.9 & 25.1 
\end{tabular}

(a) BLEU
\end{minipage}
\hfill
\begin{minipage}[t]{0.49\textwidth}
\centering
\begin{tabular}{l c c c c c}
& 5\% & 10\% & 20\% & 40\% & 100\% \\
\hline\hline
& \multicolumn{5}{c}{En-Fr} \\
\hline
NMT & 82.0 & 84.0 & 85.0 & 85.9 & 86.9 \\
LC-NMT & 82.4 & 84.8 & 85.6 & 86.0 & 86.4 \\
\hline\hline
& \multicolumn{5}{c}{En-De} \\
\hline
NMT & 76.6 & 78.9 & 80.4 & 81.4 & 81.7 \\
LC-NMT & 77.3 & 79.5 & 80.6 & 81.5 & 81.7 
\end{tabular}

(b) RIBES
\end{minipage}

\vspace{-3mm}
\caption{Translation quality in (a) BLEU and (b) RIBES on the cross-lingual pronoun prediction corpora}
\label{tab:result-quality}
\end{table*}

\begin{table}[t]
\centering
\small
\begin{tabular}{c c c c c | c}
5\% & 10\% & 20\% & 40\% & 100\% & 100\%\\
\hline\hline
\multicolumn{6}{c}{En-Fr} \\
\hline
49.7 & 54.1 & 57.6 & 64.2 & 67.6 & 65.7$^\star$ \\
50.7 & 54.0 & 60.4 & 64.2 & 59.2 & 65.35$^\circ$ \\
\hline\hline
\multicolumn{6}{c}{En-De} \\
\hline
44.6 & 44.1 & 44.9 & 50.2 & 56.4 & 64.6$^\star$ \\
54.2 & 46.3 & 44.8 & 52.3 & 51.1 & 52.5$^\bullet$
\end{tabular}

\caption{Macro-average recall for cross-lingual pronoun prediction. We display two top rankers from the shared task in the last column. ($\star$) \citep{luotolahti-kanerva-ginter:2016:WMT} ($\circ$) \citep{stymne:2016:WMT}
($\bullet$) \citep{dabre-EtAl:2016:WMT}}
\label{tab:result-pronoun}
\end{table}

\begin{table}
\centering
\begin{tabular}{l c c}
 & BLEU & RIBES\\
\hline
\hline
NMT & 19.7 & 77.8\\
LC-NMT & 20.7 & 79.0\\
\end{tabular}
\caption{\label{tab:result-quality-iwslt} Translation quality on IWSLT (En-De).}
\end{table}

\subsection{Experiments}

\subsubsection{Models and Learning}

\paragraph{Naive Model (NMT)}

We train a naive attention-based neural machine translation system based on the code publicly available online.\footnote{
    \url{https://github.com/nyu-dl/dl4mt-tutorial/}
}
The dimensionalities of word vectors, encoder recurrent network and decoder recurrent network are 620, 1000 and 1000, respectively. We use a one-layer feedforward network with one $\tanh$ hidden units as an attention model. We regularize the models with Dropout\cite{pham2014dropout}.

\paragraph{Larger-Context Model (LC-NMT)}

A larger-context model closely follows the configuration of the naive model. The additional encoder has two GRU's, and thus outputs a 2000-dimensional time-dependent context vector each time.

\paragraph{Learning}

We train both types of models to maximize the log-likelihood given a training corpus using Adadelta \cite{zeiler2012adadelta}. We early-stop with BLEU on a validation set.\footnote{We use greedy decoding for early-stopping.}
 We do not do anything particular for the cross-lingual pronoun prediction task.

\paragraph{Varying training corpus sizes}

We experiment by varying the size of the training corpus to see if there is any meaningful difference in performance between the vanilla and larger-context models w.r.t. the size of training set. We do it for the corpora from the pronoun prediction task, using 
5\%, 10\%, 20\%, 40\% and 100\% of the original training set.

\subsubsection{Results}


From the results presented in Table~\ref{tab:result-pronoun}, we observe that the larger-context models generally outperform the vanilla ones in terms of BLEU, RIBES and macro-average recall. However, this improvement vanishes as the size of training set grows. We confirm that this is not due to the lemmatization of the target side of the pronoun task corpora by observing that the proposed larger-context model also outperforms the vanilla one on IWSLT En-De, of which the training corpus size is approximately 10\% of the full pronoun task corpus, as shown in Table~\ref{tab:result-quality-iwslt}). 

\section{Conclusion}

In this paper, we have proposed a novel extension of attention-based neural machine translation that seamlessly incorporates the context from surrounding sentences. Our extensive evaluation, measured both in terms of average translation quality and cross-lingual pronoun prediction, has revealed that the benefit from larger context is moderate when there were a few training sentence pairs. We were not able to observe a similar level of benefit with a larger training corpus. We suspect that a large corpus allows the model to capture subtle word relations from a source sentence alone. We believe that a better more-focused evaluation metric may be necessary in order to properly evaluate the influence of discourse-level information in translation.

\section*{Acknowledgments}

This work was supported by Samsung Electronics (Larger-Context Neural Machine Translation). KC thanks Google (Faculty Award 2016), NVIDIA (NVAIL), Facebook and eBay for their generous support. 


\bibliography{emnlp2017}
\bibliographystyle{emnlp_natbib}

\end{document}